\documentclass[10pt,twocolumn,letterpaper]{article}

\usepackage[utf8]{inputenc}
\usepackage[T1]{fontenc}
\usepackage{times}
\usepackage[margin=0.75in,top=0.9in,bottom=1in,columnsep=0.25in]{geometry}
\usepackage{amsmath,amssymb}
\usepackage{booktabs}
\usepackage{multirow}
\usepackage{graphicx}
\usepackage{xcolor}
\usepackage{microtype}
\usepackage[numbers,sort&compress]{natbib}
\usepackage[font=small,labelfont=bf,skip=4pt]{caption}
\usepackage{enumitem}
\usepackage{url}
\usepackage[colorlinks=true,linkcolor=black,citecolor=blue!60!black,urlcolor=blue!60!black]{hyperref}

\graphicspath{{figures/}}
\newcommand{\recall}[1]{R@#1}

\title{\LARGE\bfseries The Photo Gap: How Much Retrieval Accuracy\\Consumer Photographs Cost Zero-Shot Fashion Encoders}

\author{%
  Reem Mohanty\\
  Glo AI Inc.\ (Bannabre), New York, NY
}
\date{November 2026}

\begin{document}
\maketitle

\begin{abstract}
Matching a garment to its catalog entry from a photograph is a routine component of visual search, styling and resale products, and in practice it is built on an image encoder used as released, without task-specific training. Such systems are tuned and demonstrated on catalog imagery, yet the photographs they receive from users are taken on phones, in homes, by people. We measure what that difference costs. Seven publicly released encoders (CLIP ViT-B/32, CLIP ViT-L/14, SigLIP ViT-B/16, FashionCLIP, Marqo-FashionCLIP, Marqo-FashionSigLIP and DINOv2) are evaluated zero-shot on the DeepFashion Consumer-to-Shop benchmark: 47,434 consumer photographs queried against 22,669 catalog images of 16,941 items. Recall@10 ranges from 7.9\% to 46.6\%. To separate the difficulty of the photograph from the difficulty of the item, we add a control in which catalog images of the same 4,152 garments serve as queries against the same gallery. On identical items the best encoder reaches 75.7\% Recall@10 from a catalog query and 50.5\% from a consumer query; we call the 25-point difference the \emph{photo gap}. For the other vision-language encoders the gap is 29--40 points, and for DINOv2 it is 45. Three further observations order the options available to a practitioner: the pretraining recipe of the base model is worth as much as fashion fine-tuning (a general SigLIP ViT-B/16 matches the fashion-tuned Marqo-FashionCLIP); model scale buys little (CLIP ViT-L/14 improves on ViT-B/32 by 1.7 points of Recall@1); cropping to the garment recovers about a quarter of the gap; and lower-body garments are retrieved at half the rate of tops and dresses by every model. Code reproducing every figure is released.
\end{abstract}

\section{Introduction}

Visual search for clothing has a long history in computer vision and a short one in products. The task is to find, among the images of a retailer's catalog, the item shown in a photograph that a shopper took. Street-to-shop retrieval was formulated more than a decade ago~\cite{liu2012street,kiapour2015where,liu2016deepfashion}, and the benchmarks built then remain the standard way of measuring it. What has changed is how the task is solved in practice. The systems that now answer ``find this garment'' queries in marketplace apps, styling assistants and secondhand platforms rarely train a retrieval model on consumer--catalog pairs. They embed images with a vision-language encoder released by somebody else, CLIP~\cite{radford2021clip} or SigLIP~\cite{zhai2023siglip} or a fashion-tuned derivative such as FashionCLIP~\cite{chia2022fashionclip}, and rank catalog images by cosine similarity. The encoder is used zero-shot because labelled pairs at the scale supervised methods require are expensive, and because the released encoders are good enough to ship.

How good they are on the photographs that users actually submit is less clear than it should be. The encoders are evaluated and demonstrated on catalog imagery: product pages, studio shots, flat lays photographed under controlled light. Recent benchmarks of fashion retrieval, LookBench among them~\cite{gao2026lookbench}, draw their queries from product pages, street-style photography and generated images, and score a retrieved item as relevant when it shares the query's category and attributes rather than when it is the same garment. The supervised literature on consumer-to-shop retrieval reports numbers on consumer queries, but for models trained on the benchmark's own pairs, which answers a different question. Between the two lies the quantity a product team needs: how much accuracy an off-the-shelf encoder loses when the query stops being a catalog image and becomes a photograph taken on a phone, of a garment being worn or lying on a bed, by someone who is not a photographer.

That quantity matters most where every query is such a photograph. Secondhand fashion is the clearest case: a listing on a resale marketplace is, by construction, a consumer photo, and the segment is growing roughly twice as fast as apparel as a whole~\cite{thredup2026}. A system that finds the catalog twin of a resale listing with 50\% recall and one that does so with 75\% recall are different products, and the difference between them is a property of the encoder on consumer photographs, not on catalog images.

This paper measures that property directly. We take seven publicly released image encoders, three general-purpose vision-language models, three fashion-specialised ones and one self-supervised vision-only model, and evaluate them without any fine-tuning on the DeepFashion Consumer-to-Shop benchmark~\cite{liu2016deepfashion}, using consumer photographs as queries against a gallery of catalog images. The consumer-to-shop recall of a model, taken on its own, conflates two difficulties: the photograph, and the item. A gallery of seventeen thousand garments contains many near-duplicates, and some of a model's failures would occur with a perfect photograph. To separate the two we introduce a control that holds the items fixed and changes only the query: for every garment with more than one catalog image, each catalog image is used as a query against the same gallery, so that catalog-to-catalog and consumer-to-catalog recall can be compared on identical items. The difference between them is the cost of the photograph. We call it the photo gap.

The contributions are as follows.
\begin{enumerate}[leftmargin=*,itemsep=1pt,topsep=2pt]
  \item A zero-shot benchmark of seven current image encoders on consumer-to-shop clothing retrieval, with code that reproduces every number. \recall{1} ranges from 3.7\% (CLIP ViT-B/32) to 22.2\% (Marqo-FashionSigLIP) and \recall{10} from 7.9\% to 46.6\%.
  \item The paired photo gap: the recall lost when the query is a consumer photograph rather than a catalog image, measured on identical items against an identical gallery. At \recall{10} it is 25 points for the best encoder and 29--45 points for the others; a consumer photograph retains between 18\% and 67\% of the recall the same encoder achieves from a catalog photograph of the same garment.
  \item An ordering of the levers available to a deployed system, from a crop ablation and breakdowns by garment type. The pretraining recipe of the base model matters as much as fashion fine-tuning; model scale matters little; cropping to the garment bounding box recovers about a quarter of the gap; and lower-body garments are retrieved at half the rate of upper-body and full-body garments by every model.
\end{enumerate}

\section{Related Work}

\paragraph{Cross-domain clothing retrieval.}
The problem of matching clothing photographed in uncontrolled conditions to the same item in a catalog was posed by Liu et al.~\cite{liu2012street}, who aligned body parts across the two domains, and by Huang et al.~\cite{huang2015darn}, who learned a dual attribute-aware ranking network on 450,000 street--shop pairs. Kiapour et al.~\cite{kiapour2015where} released Exact Street2Shop, with 39,479 exact matches between street and shop photographs, and showed that a learned cross-domain similarity outperforms generic deep features. Liu et al.~\cite{liu2016deepfashion} released DeepFashion, whose Consumer-to-Shop Clothes Retrieval benchmark (239,557 images, 195,540 consumer--shop pairs, 33,881 items) is the dataset used here, together with FashionNet, trained jointly on categories, attributes and landmarks. DeepFashion2~\cite{ge2019deepfashion2} scaled the setting to 873,000 commercial--consumer pairs with dense annotations. Methods trained on the DeepFashion training split have improved steadily: in the comparison assembled by Kuang et al.~\cite{kuang2019grnet}, top-1 accuracy rises from 7.0\% for FashionNet through 13.7\% for an attention model~\cite{wang2017vam}, 18.6\% for randomized ensembles~\cite{xuan2018dreml} and 21.3\% for Kronecker-product matching~\cite{shen2018kpm} to 25.7\% for their graph reasoning network, with top-20 accuracy rising from 18.8\% to 64.4\%. Every method in that line is trained on the benchmark's own pairs. We ask the complementary question, how far encoders that have never seen such pairs get, which to our knowledge has not been reported for current models.

\paragraph{General vision-language encoders.}
CLIP~\cite{radford2021clip} established that contrastive image--text pretraining on web-scale data produces image embeddings that transfer zero-shot to classification and retrieval. SigLIP~\cite{zhai2023siglip} replaced the softmax contrastive objective with a pairwise sigmoid loss and improved zero-shot accuracy at a fixed compute budget; SigLIP~2~\cite{tschannen2025siglip2} added captioning and self-supervised objectives. Because these models are released with weights and need no labels, they have become the default image encoder in deployed visual search, which is why we study them rather than task-trained models. DINOv2~\cite{oquab2024dinov2}, a self-supervised vision-only encoder, is included to separate the contribution of language supervision from that of large-scale pretraining.

\paragraph{Fashion-specific encoders.}
FashionCLIP~\cite{chia2022fashionclip} fine-tuned CLIP ViT-B/32 on roughly 800,000 Farfetch product image--text pairs and reported gains over CLIP on fashion classification and retrieval tasks; the released checkpoint starts from a LAION-2B model. Marqo-FashionCLIP and Marqo-FashionSigLIP~\cite{marqo2024fashion} fine-tune CLIP and SigLIP backbones with multi-part text supervision over product titles, categories and attributes, and report improvements over FashionCLIP across several fashion retrieval datasets. All three are trained on catalog imagery. Whether the fashion-specific gain survives a consumer photograph is one of the questions addressed here.

\paragraph{Fashion retrieval benchmarks.}
Fashion200K~\cite{han2017fashion200k} and the DeepFashion In-Shop benchmark~\cite{liu2016deepfashion} evaluate catalog-to-catalog retrieval. LookBench~\cite{gao2026lookbench} is a live, time-stamped benchmark assembled from current product pages, street-style photography and generated images, designed to resist training-set contamination. Its queries are studio packshots and street-style outfit photographs collected through an image search engine rather than photographs taken by shoppers, its relevance labels are defined by garment category and attributes rather than by item identity, and its authors report that strong baselines remain below 60\% fine \recall{1}. The present study asks a narrower question on a fixed public dataset: for the identical garment, how much recall does a shopper's photograph cost, with catalog-to-catalog retrieval on the same items as the control.

\paragraph{Domain gap.}
The street-to-shop literature has from the start attributed its difficulty to the gap between uncontrolled consumer photographs and controlled studio photographs~\cite{liu2012street,kiapour2015where}. We adopt that framing and measure the gap directly for zero-shot encoders, holding the items fixed, which the standard consumer-to-shop metric does not do.

\section{Benchmark and Protocol}
\label{sec:protocol}

\subsection{Data}
We use the Consumer-to-Shop Clothes Retrieval benchmark of DeepFashion~\cite{liu2016deepfashion}. It pairs photographs taken by shoppers (``consumer'' images: a garment being worn or laid out, photographed on a phone or point-and-shoot camera) with the retailer's photographs of the same garment (``shop'' images). The release contains 239,557 images of 33,881 distinct items (194,165 consumer and 45,392 shop images), each annotated with a clothing type (upper-body, lower-body, full-body), a category and a garment bounding box. The authors' partition assigns 120,927 images to training, 59,884 to validation and 58,746 to test. We do not use the training split. Every model is evaluated as released by its authors.

Following the benchmark's own definition of the task, the query set consists of the consumer images whose pair falls in the test split (47,434 queries), and the gallery consists of the shop images in the validation and test splits (22,669 images of 16,941 items). A query is scored as a hit at rank $K$ if any of the $K$ highest-scoring gallery images shares its item identifier. The dataset is released for non-commercial research and was obtained under the authors' agreement; it was used only for this evaluation, and no image from it is reproduced here.

\subsection{Models}
Seven publicly released image encoders, in three families, are evaluated:
\begin{itemize}[leftmargin=*,itemsep=1pt,topsep=2pt]
  \item \emph{General-purpose vision-language models:} CLIP ViT-B/32 and CLIP ViT-L/14~\cite{radford2021clip} (OpenAI weights), and SigLIP ViT-B/16 trained on WebLI~\cite{zhai2023siglip}.
  \item \emph{Fashion-specialised vision-language models:} FashionCLIP~\cite{chia2022fashionclip}, Marqo-FashionCLIP and Marqo-FashionSigLIP~\cite{marqo2024fashion}, fine-tuned from CLIP or SigLIP checkpoints on fashion product data.
  \item \emph{Vision-only self-supervised model:} DINOv2 ViT-B/14~\cite{oquab2024dinov2}, using the CLS token.
\end{itemize}
Each model is run with the preprocessing distributed with its weights, through OpenCLIP~\cite{ilharco2021openclip} for the CLIP, SigLIP and Marqo models and through Hugging Face Transformers~\cite{wolf2020transformers} for FashionCLIP and DINOv2. Image embeddings are $\ell_2$-normalised and compared by cosine similarity. No text is used at any point; the text towers are not loaded.

\subsection{Cropping}
The benchmark supplies a garment bounding box for every image. In the main setting each image is cropped to its box, enlarged by 10\% of the box width and height on every side, before the model's own preprocessing. This removes most of the background, body and clutter that distinguish a consumer photograph from a catalog one, and is therefore the setting most favourable to the consumer query. A second setting feeds the full image. Both settings are run for CLIP ViT-B/32 and for Marqo-FashionSigLIP, the weakest and the strongest encoder, so that the effect of framing can be separated from the effect of the photograph itself.

\subsection{Metrics}
We report Recall@$K$ for $K \in \{1, 5, 10, 20, 50\}$ and the mean reciprocal rank (MRR) of the first correct gallery image. Recall@$K$ is the fraction of queries for which a correct item appears among the top $K$; it is the quantity a user of a ``find this garment'' feature experiences, namely whether the right listing is on the first screen.

\subsection{The control: catalog queries on the same items}
\label{sec:control}
A consumer-to-shop recall figure cannot by itself say how much of the loss is due to the photograph and how much to the difficulty of distinguishing seventeen thousand similar garments. We therefore add a control that holds the items fixed and changes only the query. For every item with at least two shop images (4,152 items, 9,880 shop images), each shop image is used in turn as a query against the full gallery with itself removed; a hit is another shop image of the same item. This is retrieval between two catalog photographs of one garment, usually different views such as front, back and detail, scored against the same gallery with the same metrics.

We then restrict the consumer-to-shop queries to those 4,152 items (20,000 consumer queries; the \emph{paired} consumer-to-shop score) and report
\begin{align}
  \mathrm{gap}@K \;=\;{}& \recall{K}(\text{shop}\to\text{shop}) \notag\\
  & -\; \recall{K}(\text{consumer}\to\text{shop},\ \text{same items}).
\end{align}
Both terms are computed on identical items against an identical gallery with the same encoder, so the difference isolates what the consumer photograph costs. Because the catalog queries in the control are themselves different views of the garment rather than duplicates, the control is not a trivial ceiling, and the gap is a conservative estimate of that cost.

\subsection{Breakdowns}
Consumer-to-shop recall is additionally reported by clothing type (26,166 upper-body, 11,994 lower-body and 9,274 full-body queries) and, for the best encoder, by category for categories with at least 50 test queries.

\subsection{Implementation}
All runs use one NVIDIA T4 GPU in Google Colab, batch size 256, single-precision inference, PyTorch 2.14. Embedding the 70,103 query and gallery images takes between four minutes (CLIP ViT-B/32, about 290 images per second, where JPEG decoding rather than the GPU is the bottleneck) and twelve minutes (CLIP ViT-L/14, about 100 images per second) per model; no image failed to decode. Retrieval is exact: the full cosine-similarity matrix is computed in chunks of 2,048 queries and the top 50 selected by partial sort, so no number carries approximate-search error. Zero-shot inference with exact retrieval is deterministic. With 47,434 queries the binomial 95\% interval around a recall value is about $\pm 0.45$ points, so differences of one point or more between models in the tables below are not attributable to sampling. Code, including a synthetic-data test that exercises the pipeline end to end, is available at \url{https://github.com/reemmohanty16/resale-retrieval-bench}.

\section{Results}
\label{sec:results}

All figures are percentages over the 47,434 test consumer queries against the 22,669-image gallery, with the garment-box crop unless stated otherwise.

\subsection{Zero-shot consumer-to-shop retrieval}

\begin{table}[t]
\centering\small
\caption{Consumer$\to$shop retrieval, zero-shot, garment-box crop. 47,434 queries, 22,669-image gallery. Best value per column in bold.}
\label{tab:main}
\setlength{\tabcolsep}{2.6pt}
\begin{tabular}{@{}lrrrrrr@{}}
\toprule
Model & \recall{1} & \recall{5} & \recall{10} & \recall{20} & \recall{50} & MRR \\
\midrule
\multicolumn{7}{@{}l}{\emph{General vision-language}}\\
CLIP ViT-B/32        & 3.7 & 6.5 & 7.9 & 9.8 & 13.1 & .051 \\
CLIP ViT-L/14        & 5.4 & 9.1 & 10.9 & 13.1 & 17.0 & .073 \\
SigLIP ViT-B/16      & 15.6 & 29.5 & 35.7 & 42.2 & 51.5 & .223 \\
\midrule
\multicolumn{7}{@{}l}{\emph{Fashion-specialised}}\\
FashionCLIP          & 8.7 & 16.6 & 20.6 & 25.2 & 32.6 & .127 \\
Marqo-FashionCLIP    & 15.5 & 29.2 & 35.7 & 42.6 & 52.5 & .222 \\
Marqo-FashionSigLIP  & \textbf{22.2} & \textbf{39.7} & \textbf{46.6} & \textbf{53.4} & \textbf{62.2} & \textbf{.304} \\
\midrule
\multicolumn{7}{@{}l}{\emph{Vision-only}}\\
DINOv2 ViT-B/14      & 4.0 & 6.7 & 8.3 & 10.2 & 13.8 & .054 \\
\bottomrule
\end{tabular}
\end{table}

\begin{figure}[t]
\centering
\includegraphics[width=\linewidth]{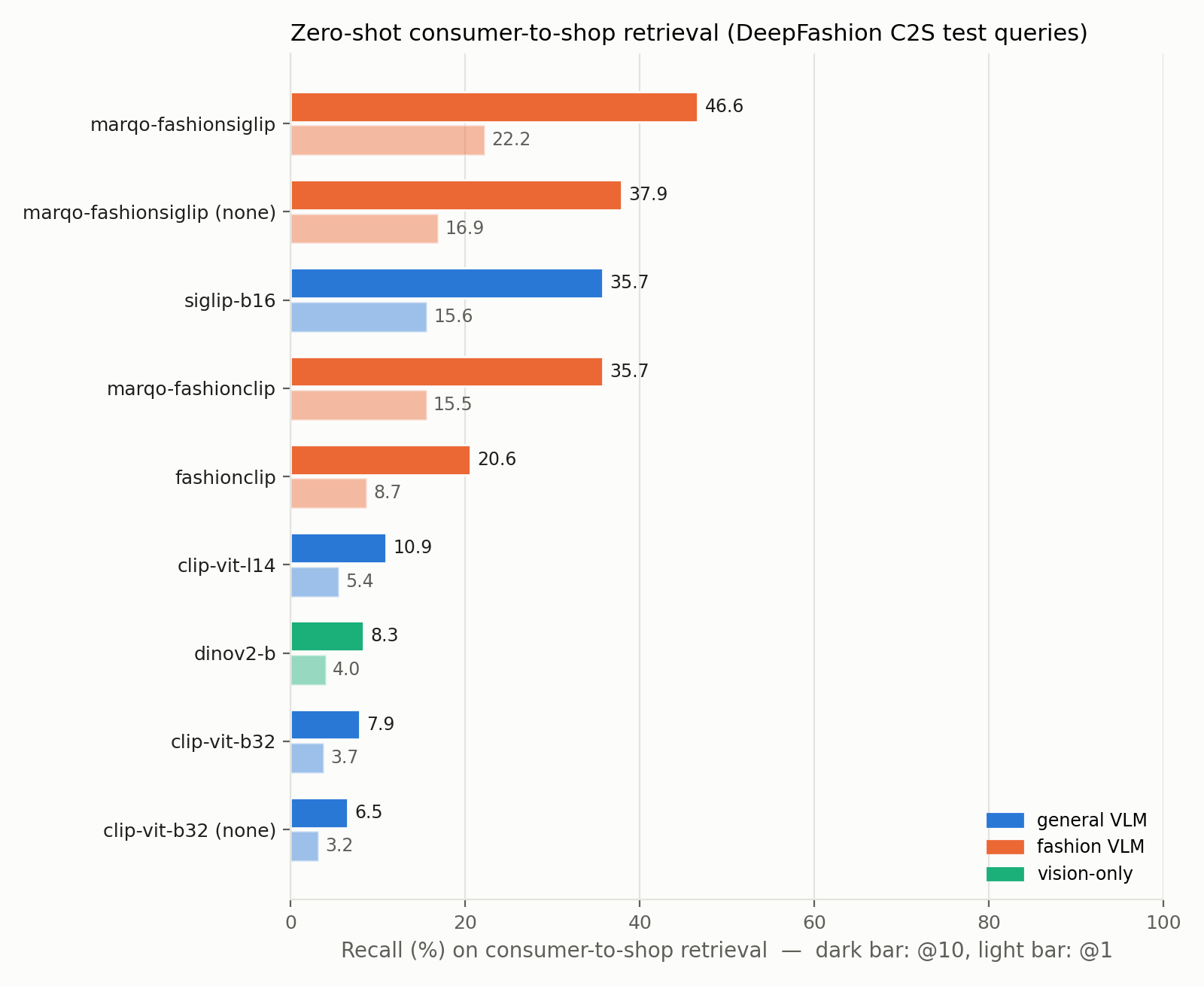}
\caption{Zero-shot consumer$\to$shop recall by encoder. Light bars: \recall{1}; dark bars: \recall{10}. Blue: general vision-language models; orange: fashion-specialised; green: vision-only. Rows marked ``(none)'' are the full-image runs of the crop ablation (Section~\ref{sec:crop}).}
\label{fig:main}
\end{figure}

Table~\ref{tab:main} and Figure~\ref{fig:main} give the main result. The spread between encoders is sixfold. The strongest, Marqo-FashionSigLIP, places the exact catalog item among the top ten for 46.6\% of consumer photographs and among the top twenty for 53.4\%; the weakest, CLIP ViT-B/32, does so for 7.9\% and 9.8\%. Three patterns in the table bear on how such a system should be built.

\emph{Fashion fine-tuning helps, and the base model it starts from helps as much.} Fine-tuning CLIP ViT-B/32 on fashion data raises \recall{1} from 3.7 to 8.7 (FashionCLIP) and to 15.5 (Marqo-FashionCLIP). A general-purpose SigLIP ViT-B/16 that has seen no fashion-specific data reaches 15.6, indistinguishable from Marqo-FashionCLIP. Fine-tuning SigLIP on fashion data (Marqo-FashionSigLIP) adds a further seven points, to 22.2. The pretraining recipe of the base model is worth about as much as the domain data placed on top of it.

\emph{Scale does not help.} CLIP ViT-L/14 has roughly four times the compute of ViT-B/32 and gains 1.7 points of \recall{1} (3.7 to 5.4) and 3.0 points of \recall{10}. A SigLIP ViT-B/16 of a third its size triples its recall. On consumer photographs, what a model was trained on matters far more than how large it is.

\emph{Vision-only self-supervision does not transfer.} DINOv2 ViT-B/14, a strong general-purpose feature extractor, scores 4.0, level with the weakest CLIP. Without either language supervision or fashion data, features that serve dense and category-level tasks well do not pick out the identity of a garment across a change of photograph.

For scale, the supervised methods summarised in Section~2 report top-1 accuracy between 7.0\% (FashionNet, 2016) and 25.7\% (GRNet, 2019), and top-20 accuracy between 18.8\% and 64.4\%, against a gallery of all 45,392 shop images, twice the size of ours. Read with that caveat, a zero-shot Marqo-FashionSigLIP (22.2 / 53.4) falls within the range of supervised systems of 2018--2019 without having seen a single consumer--shop training pair, while a zero-shot CLIP falls below the 2016 baseline.

\subsection{The photo gap}

\begin{table*}[t]
\centering\small
\caption{The photo gap. Shop$\to$shop (control) and consumer$\to$shop recall on the same 4,152 items (9,880 shop queries, 20,000 consumer queries), garment-box crop. ``Retained'' is the consumer$\to$shop \recall{10} as a fraction of the shop$\to$shop \recall{10}. Smallest gap and largest retained fraction in bold.}
\label{tab:gap}
\begin{tabular}{@{}lrrrrrrr@{}}
\toprule
 & \multicolumn{3}{c}{Recall@1} & \multicolumn{3}{c}{Recall@10} & \\
\cmidrule(lr){2-4}\cmidrule(lr){5-7}
Model & shop$\to$shop & consumer$\to$shop & gap & shop$\to$shop & consumer$\to$shop & gap & Retained \\
\midrule
CLIP ViT-B/32       & 23.5 & 4.7  & 18.9 & 45.8 & 10.2 & 35.7 & 22\% \\
CLIP ViT-L/14       & 26.7 & 6.9  & 19.8 & 49.3 & 14.1 & 35.2 & 29\% \\
SigLIP ViT-B/16     & 42.0 & 17.6 & 24.4 & 72.4 & 39.8 & 32.6 & 55\% \\
FashionCLIP         & 35.2 & 10.4 & 24.8 & 63.7 & 24.0 & 39.7 & 38\% \\
Marqo-FashionCLIP   & 36.2 & 17.8 & 18.4 & 68.9 & 40.0 & 28.9 & 58\% \\
Marqo-FashionSigLIP & 39.8 & 24.4 & \textbf{15.3} & 75.7 & 50.5 & \textbf{25.2} & \textbf{67\%} \\
DINOv2 ViT-B/14     & 33.0 & 4.9  & 28.1 & 55.7 & 10.2 & 45.4 & 18\% \\
\bottomrule
\end{tabular}
\end{table*}

\begin{figure}[t]
\centering
\includegraphics[width=\linewidth]{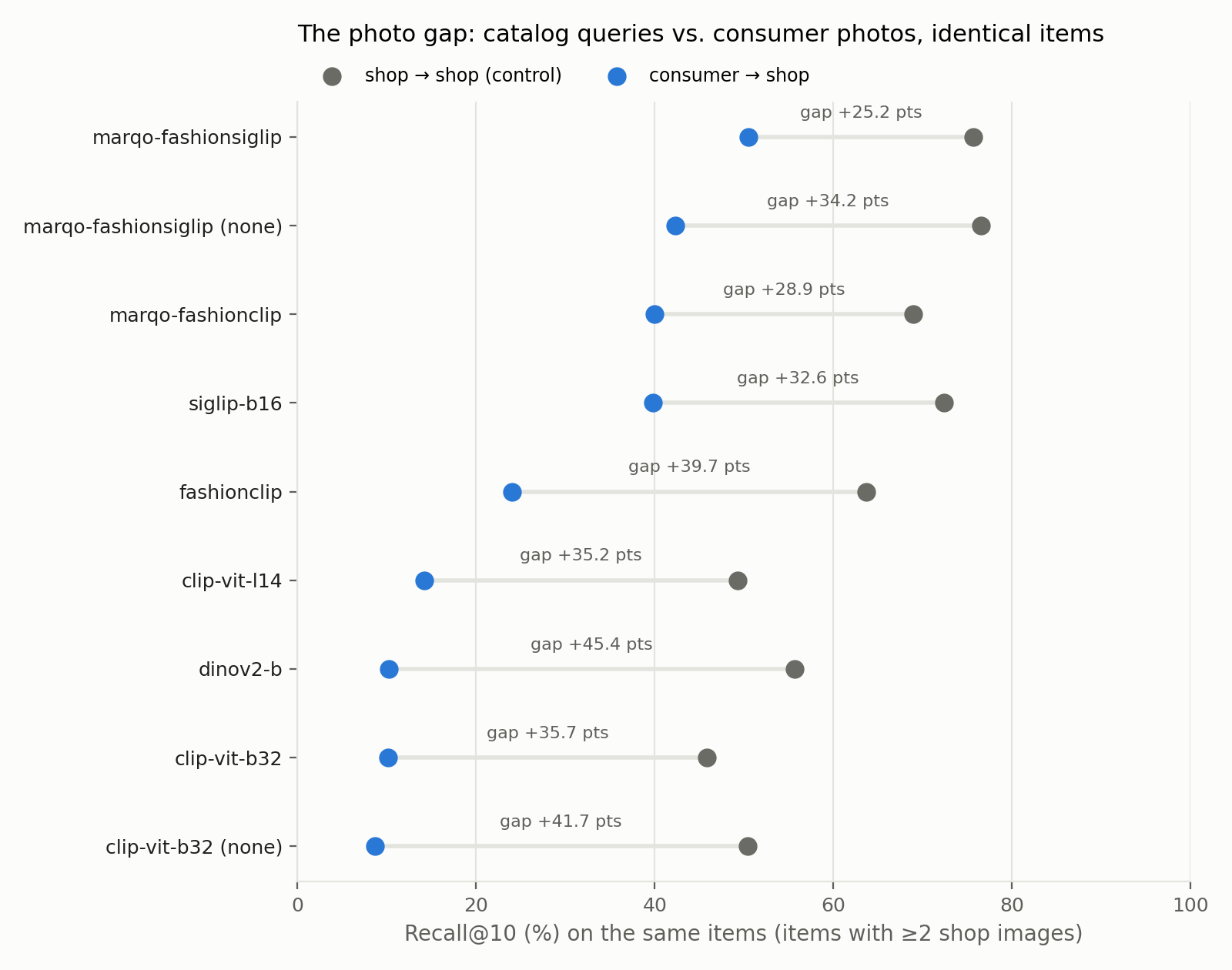}
\caption{The photo gap at \recall{10}. For each encoder, recall when the query is another catalog photograph of the garment (grey) and when it is a consumer photograph of the same garment (blue), on the same 4,152 items. The label gives the difference.}
\label{fig:gap}
\end{figure}

Table~\ref{tab:gap} and Figure~\ref{fig:gap} report the control described in Section~\ref{sec:control}. The control is not a trivial ceiling: even the best encoder finds another catalog view of the same garment among the top ten only 75.7\% of the time, because the shop images of an item are different views of it. The gap is therefore the cost of the consumer photograph over and above a change of catalog view.

That cost is large for every encoder. At \recall{10} the vision-language models lose between 25 and 40 points and DINOv2 loses 45. Replacing the catalog photograph of a garment with a shopper's photograph of the same garment removes a third of the best model's recall and three quarters of CLIP's.

Two features of the table are not visible in the headline ranking. First, the ordering by retained fraction differs from the ordering by absolute score. DINOv2 has a higher shop-to-shop score than CLIP ViT-L/14 (55.7 against 49.3) and the lowest consumer-to-shop score of all; of the seven encoders its features are the most sensitive to the change of photograph. Second, fashion fine-tuning narrows the gap rather than merely raising the ceiling. From SigLIP to Marqo-FashionSigLIP the shop-to-shop score rises by 3 points at \recall{10} and the consumer-to-shop score by 11; from CLIP ViT-B/32 to Marqo-FashionCLIP, by 23 and 30. In both pairs the consumer side gains more than the catalog side, although the fine-tuning data is catalog imagery.

\subsection{Cropping to the garment}
\label{sec:crop}

\begin{table}[t]
\centering\small
\caption{Full image versus garment-box crop (10\% padding) for the weakest and the strongest encoder; same gallery and protocol. ``Paired'' is consumer$\to$shop recall on the 4,152 control items; the gap is shop$\to$shop minus paired.}
\label{tab:crop}
\footnotesize
\setlength{\tabcolsep}{2.2pt}
\begin{tabular}{@{}llrrrrr@{}}
\toprule
 & & \multicolumn{2}{c}{cons.$\to$shop} & shop$\to$shop & paired & \\
\cmidrule(lr){3-4}
Model & Input & \recall{1} & \recall{10} & \recall{10} & \recall{10} & gap@10 \\
\midrule
CLIP B/32 & full  & 3.2 & 6.5 & 50.4 & 8.6 & 41.7 \\
CLIP B/32 & crop  & 3.7 & 7.9 & 45.8 & 10.2 & 35.7 \\
\addlinespace
Marqo-FashionSigLIP & full & 16.9 & 37.9 & 76.6 & 42.3 & 34.2 \\
Marqo-FashionSigLIP & crop & 22.2 & 46.6 & 75.7 & 50.5 & 25.2 \\
\bottomrule
\end{tabular}
\end{table}

Table~\ref{tab:crop} compares full-image and cropped inputs. Cropping raises consumer-to-shop recall (by 5.3 points of \recall{1} and 8.7 of \recall{10} for Marqo-FashionSigLIP) and slightly lowers shop-to-shop recall (by 0.9 to 4.6 points: catalog images are already clean, and the crop discards a little context). The gap shrinks by 6 points for CLIP and by 9 for Marqo-FashionSigLIP. About a quarter of the photo gap, then, is framing and background, which a detector can remove at inference time. The remaining 25 points for the best encoder lie in the garment as photographed, in its lighting, pose, creasing, partial occlusion by the body and viewpoint, rather than in what surrounds it.

\subsection{Which garments}

\begin{table}[t]
\centering\small
\caption{Consumer$\to$shop \recall{10} by clothing type, garment-box crop (26,166 upper-body, 11,994 lower-body and 9,274 full-body queries).}
\label{tab:type}
\begin{tabular}{@{}lrrr@{}}
\toprule
Model & Upper-body & Lower-body & Full-body \\
\midrule
CLIP ViT-B/32       & 9.7  & 5.6  & 5.7 \\
CLIP ViT-L/14       & 13.6 & 7.2  & 8.4 \\
SigLIP ViT-B/16     & 40.1 & 22.1 & 41.0 \\
FashionCLIP         & 23.6 & 13.3 & 21.4 \\
Marqo-FashionCLIP   & 41.4 & 19.9 & 40.0 \\
Marqo-FashionSigLIP & 52.5 & 27.1 & 55.5 \\
DINOv2 ViT-B/14     & 9.9  & 6.1  & 6.8 \\
\bottomrule
\end{tabular}
\end{table}

\begin{figure}[t]
\centering
\includegraphics[width=0.92\linewidth]{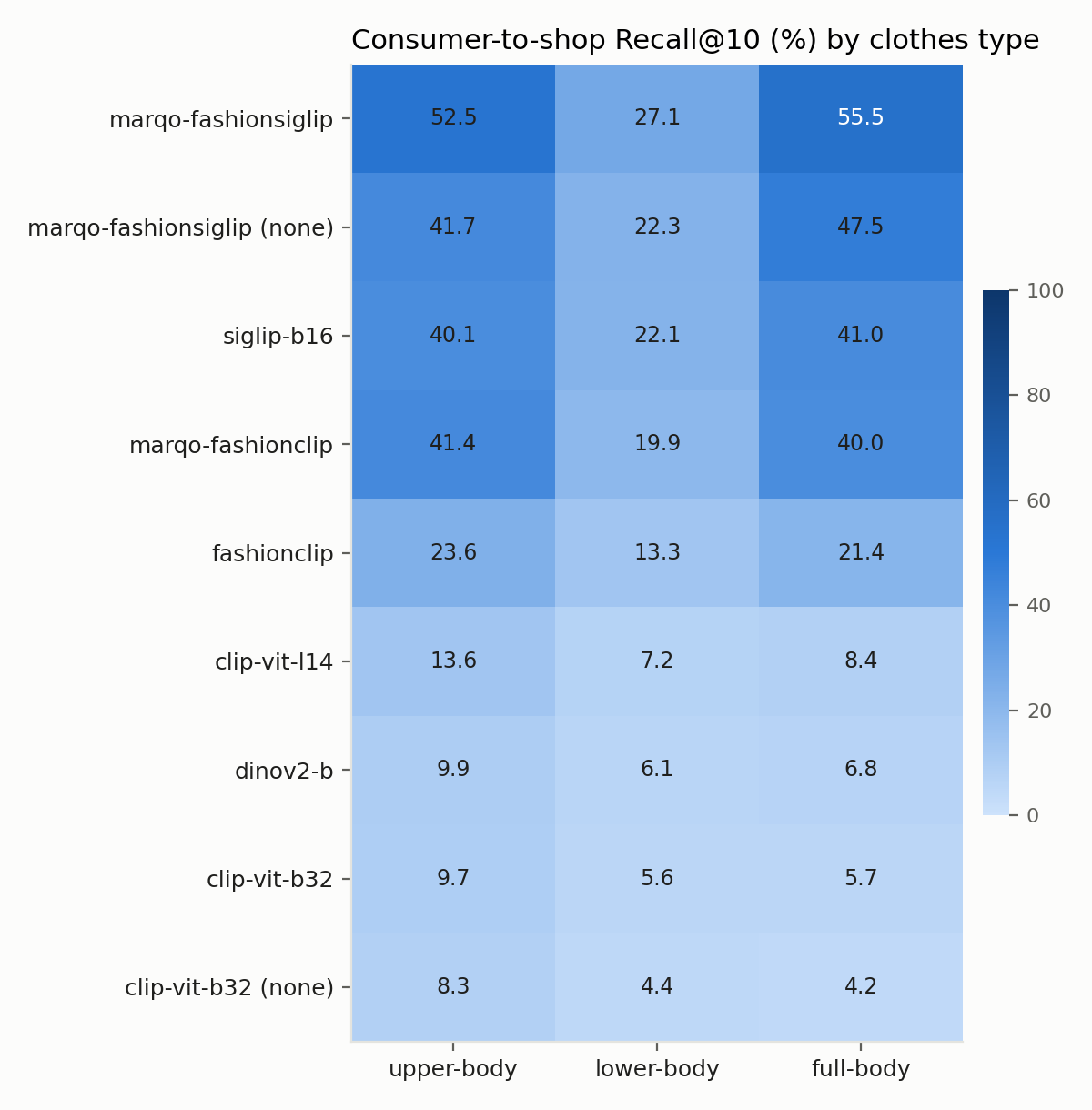}
\caption{Consumer$\to$shop \recall{10} by clothing type and encoder. Lower-body garments are the difficult column for every encoder.}
\label{fig:type}
\end{figure}

Table~\ref{tab:type} and Figure~\ref{fig:type} break the result down by clothing type. Lower-body garments are retrieved at about half the rate of upper-body and full-body garments by every encoder, including the fashion-tuned ones. By category (Appendix~\ref{app:category}), Marqo-FashionSigLIP reaches 69.8\% \recall{10} on polo shirts, 59.8\% on dresses and 59.1\% on T-shirts, but 32.4\% on trousers, 28.4\% on jeans and 20.9\% on leggings. The pattern is consistent with how people photograph bottoms: from above, by the wearer, cropped at the waist, with the legs bent, whereas the catalog shows them flat or on a standing model. Bottoms also carry less print and texture than tops, so their identity rests on silhouette and fit, which is precisely what the consumer viewpoint distorts.

\section{Discussion}

\paragraph{What a product built on these encoders can promise.}
With the best publicly available encoder and a garment crop, a shopper's photograph finds its exact catalog counterpart among the top ten results about half the time (46.6\%) and among the top twenty slightly more than half (53.4\%); for lower-body garments, about one time in four. That is a useful browsing aid and well short of search in the sense a user expects. The same encoder, given a catalog photograph instead, finds the counterpart three times in four. A system that lets a user photograph a garment and shows ``this exact item'' is, with current off-the-shelf encoders, showing the right item on the first page for roughly every second query, and should be designed and described accordingly: as a ranked shortlist for the user to confirm, not as a lookup.

\paragraph{Which lever to pull, and in what order.}
The results order the options open to a team without a labelled consumer--catalog dataset. The first choice is the base encoder: at equal size, a SigLIP encoder retrieves three to four times as many consumer photographs correctly as a CLIP encoder, and the choice costs nothing. The second is a fashion-tuned checkpoint of that encoder, which adds about seven points of \recall{1} on top of SigLIP and narrows the gap more than it raises the ceiling. The third is to crop to the garment: a detector is cheap at inference time and recovers about a quarter of the gap. Model scale is not on the list; the four-times-larger CLIP gained under two points.

\paragraph{Where the remaining gap lies.}
After cropping, 25 points of \recall{10} separate a catalog query from a consumer query of the same garment with the best encoder. That residue is in the pixels of the garment as worn and lit, not in the frame around it, so the remaining inexpensive interventions are on the capture and training sides rather than in preprocessing: asking for a second, flat-lay photograph and querying with both; or fine-tuning on consumer--shop pairs, which is what the supervised methods of Section~2 do and what no public checkpoint has been trained for. Either can be evaluated with the protocol used here.

\paragraph{Bottoms are a separate problem.}
Lower-body garments are retrieved at half the rate of tops by every encoder, fashion-tuned or not, so fashion fine-tuning as currently practised does not address it. The viewpoint of a self-taken photograph of trousers removes the silhouette cue on which bottoms depend. A system might treat bottoms with their own capture guidance (full length, from the front) or lean on non-visual attributes such as rise, inseam, fit and brand for that category.

\paragraph{The resale case.}
DeepFashion's consumer photographs are shopper photographs of garments bought new, taken roughly a decade ago. Secondhand listing photographs are if anything harder: worn garments, mixed lighting, hangers and floors, and no guarantee that a catalog image of the item exists at all. The gap measured here is therefore a floor for the resale setting. Measuring it on real listing photographs, collected with permission from a resale store, is the natural next step and the one we intend to take.

\section{Limitations}

The study is zero-shot only. No supervised or fine-tuned model was re-run under our protocol, and the supervised figures quoted from Kuang et al.~\cite{kuang2019grnet} use a gallery of 45,392 images, twice ours, so the comparison in Section~\ref{sec:results} is indicative rather than head-to-head.

The control is conservative. Shop-to-shop queries are different catalog views of a garment, not duplicates, so the gap understates what the consumer photograph costs relative to a same-view ceiling. The control subset (4,152 items with at least two shop images, 12\% of gallery items) is also slightly easier than the full query set: Marqo-FashionSigLIP scores 24.4 \recall{1} on it against 22.2 overall.

Training-data overlap cannot be ruled out for the fashion-tuned encoders. The Marqo model cards do not name their fine-tuning data; they list DeepFashion In-Shop and DeepFashion Multimodal among the evaluation sets, which suggests those were held out, and do not mention Consumer-to-Shop either way. FashionCLIP was trained on Farfetch catalog data. The absolute scores of the fashion-tuned models carry this caveat. The photo gap, which compares two query sets against one gallery with one encoder, is less exposed to it.

All conclusions rest on a single dataset of consumer photographs taken around 2016. The numbers are deterministic and the query set is large enough that one-point differences are real, but modern phone cameras may narrow the gap, and the cluttered photographs typical of resale listings may widen it.

The crop results use the annotated bounding boxes. A deployed system needs a detector, whose errors would reduce the gains in Section~\ref{sec:crop}; those gains are an upper bound.

Finally, the study isolates the image encoder. Production systems usually combine image similarity with text and attribute signals, which could compensate for part of the gap, particularly for bottoms.

\section{Conclusion}

Off-the-shelf image encoders are the component on which photograph-to-catalog search in fashion is built, and they are evaluated almost entirely on catalog imagery. Measured on the photographs users actually take, the best public encoder places the right item among the top ten for 46.6\% of queries, and on identical items a consumer photograph costs it 25 points of \recall{10} relative to a catalog photograph; weaker encoders lose 29--45 points. The pretraining recipe of the base model matters as much as fashion fine-tuning, model scale matters little, cropping to the garment recovers a quarter of the gap, and lower-body garments remain a problem for every model. The protocol, which holds the items fixed and varies only the photograph, is simple to apply to any encoder and any paired dataset, and the code to do so is released.

\section*{Acknowledgments}
We thank Ziwei Liu and the Multimedia Laboratory at the Chinese University of Hong Kong for access to the DeepFashion Consumer-to-Shop benchmark.

\bibliographystyle{plainnat}
\bibliography{refs}

\appendix

\section{Per-category results}
\label{app:category}

\begin{table}[h]
\centering\small
\caption{Consumer$\to$shop recall of Marqo-FashionSigLIP by category (garment-box crop), categories with at least 50 test queries, sorted by \recall{10}. Category names are the dataset's folder names.}
\label{tab:category}
\begin{tabular}{@{}lrrr@{}}
\toprule
Category & Queries & \recall{1} & \recall{10} \\
\midrule
Polo shirt   & 199    & 35.2 & 69.8 \\
Dress        & 3,811  & 30.2 & 59.8 \\
T-shirt      & 8,340  & 31.6 & 59.1 \\
Tank top     & 1,463  & 25.8 & 51.5 \\
Blouse       & 373    & 22.5 & 47.5 \\
Lace shirt   & 1,530  & 19.9 & 46.7 \\
Coat         & 3,659  & 20.0 & 45.8 \\
Skirt        & 632    & 16.9 & 45.4 \\
Chiffon      & 389    & 23.1 & 45.0 \\
Summer wear  & 21,264 & 20.4 & 44.0 \\
Pants        & 3,512  & 12.6 & 32.4 \\
Jeans        & 306    & 8.2  & 28.4 \\
Leggings     & 1,899  & 8.4  & 20.9 \\
\midrule
\multicolumn{4}{@{}l}{\emph{By top-level group}}\\
Clothing     & 2,481  & 29.6 & 55.5 \\
Dresses      & 4,500  & 28.1 & 57.5 \\
Tops         & 30,066 & 25.3 & 51.9 \\
Trousers     & 10,387 & 8.7  & 24.7 \\
\bottomrule
\end{tabular}
\end{table}

\section{Reproduction}
\label{app:repro}

The repository at \url{https://github.com/reemmohanty16/resale-retrieval-bench} contains the parsing, embedding, evaluation and reporting code, the summary table and figures of the run reported here, and a Colab notebook that reproduces it on a T4 in about an hour given access to the dataset. Embedding is invoked once per model and crop setting, writes $\ell_2$-normalised embeddings aligned with an index of image paths, item identifiers, clothing types and partitions, and is skipped if the output already exists; evaluation then computes every metric in this paper from those files. A synthetic-data test constructs a small dataset in the DeepFashion layout, checks that identical consumer and shop images give \recall{1} = 1.0, and runs every backend with randomly initialised weights, so that the pipeline can be verified without the dataset or any download. The dataset itself must be obtained from its authors under their release agreement and is not redistributed.

\end{document}